\pdfoutput=1

\documentclass[11pt]{article}

\usepackage[final]{coling}

\usepackage{times}
\usepackage{latexsym}
\usepackage{graphicx}
\usepackage{amsmath}
\usepackage{amssymb}
\usepackage{hyperref}

\usepackage[T1]{fontenc}

\usepackage[utf8]{inputenc}

\usepackage{microtype}

\usepackage{inconsolata}

\usepackage{graphicx}

\title{Extracting General-use Transformers for Low-resource Languages via Knowledge Distillation}

\author{
Jan Christian Blaise Cruz \textnormal{and} Alham Fikri Aji
\\
{MBZUAI}\quad
\\
\texttt{ \{jan.cruz,alham.fikri\}@mbzuai.ac.ae}
\\
}

\begin{document}
\maketitle
\begin{abstract}
In this paper, we propose the use of simple knowledge distillation to produce smaller and more efficient single-language transformers from Massively Multilingual Transformers (MMTs) to alleviate tradeoffs associated with the use of such in low-resource settings. Using Tagalog as a case study, we show that these smaller single-language models perform on-par with strong baselines in a variety of benchmark tasks in a much more efficient manner. Furthermore, we investigate additional steps during the distillation process that improves the soft-supervision of the target language, and provide a number of analyses and ablations to show the efficacy of the proposed method.
\end{abstract}

\section{Introduction}
To curb the detrimental effects of pretraining with very little pretraining data in a low-resource language, most works opt to use pretrained \textit{Massively Multilingual Transformers} (MMTs) such as  mBERT \cite{devlin-etal-2019-bert} and mDeBERTa \cite{he2021deberta,he2021debertav3} instead.

However, this comes with a number of tradeoffs. Finetuning in only one language causes negative interference in a model that compresses many languages within a limited parameter budget \cite{berend-2022-combating,lee-hwang-2023-multilingual}. This would mean that an MMT, in theory, would perform worse than using a transformer pretrained in one specific language \cite{cruz-cheng-2022-improving,pfeiffer-etal-2022-lifting}. Additionally, MMTs are unnecessarily costly as most researchers who use them are only interested in one language among many -- this is most especially the case in low-resource language research communities that also suffer from a lack of computational resources \cite{alabi-etal-2022-adapting,ansell-etal-2023-distilling}.

In this work, we propose the use of simple knowledge distillation to extract robust and efficient single-language pretrained transformers from an MMT. We study a number of intermediate steps that improve the distillation method, such as target-language conditioning and student initialization. We then compare the performance of our extracted models on strong baselines on a variety of benchmark tasks and perform ablations and analyses to pinpoint the sources of strong performance from our simple method.



\section{Methodology}

\subsection{Distillation}
To simplify the study, we limit ourselves to one type of MMT -- mBERT (\texttt{bert-base-multilingual-cased}) \cite{devlin-etal-2019-bert} -- and one language (Tagalog) for both distillation and task finetuning.

In the interest of resource-scarce research settings, the proposed method is \textit{very} simple and computationally cheap: we take a pretrained mBERT and freeze its weights. We then construct a blank student transformer with a modified architecture and use teacher-student model distillation \cite{hinton2015distillingknowledgeneuralnetwork} using masked language modeling (MLM) as the main objective. No further tricks, post-processing, or augmentations are done after distillation. We use OSCAR's Tagalog split \cite{OrtizSuarezSagotRomary2019} as the training corpus for knowledge distillation. 

Mathematically, we optimize our distillation loss as a mix of the weighted sum of the Kullback-Leibler (KL) divergence and the MLM loss between the student and teacher's output logits:

\begin{equation}
\begin{split}
L_{\textnormal{distil}} &= \alpha_{\textnormal{KL}} \textnormal{KL}(out_{student} || out_{teacher}) +\\
    &= \alpha_{\textnormal{MLM}} L_{\textnormal{MLM}}(out_{student}, out_{teacher})
\end{split}
\end{equation}
where $\alpha_{kl}$ and $\alpha_{mlm}$ represent the weights of the divergence and the MLM loss respectively to the final distillation loss. For our experiments, we use cross entropy as our MLM loss. Note that we also apply a temperature parameter to cool down the logits of the student and teacher and encourage diversity in outputs.

This gives us a distilled version of the pretrained mBERT but without the risk of negative interference caused by parameter sharing between multiple languages in the model during downstream finetuning. We produce two distilled models this way which we refer to as \texttt{dBERT Base} and \texttt{dBERT Tiny}, depending on the hyperparameters used. Hyperparameter choices used for distillation are listed on Table \ref{tab:distillation_parameters}. We run distillation for a total of three epochs on the training dataset.


\begin{table}[]
    \centering
    \begin{tabular}{|c|c|c|c|}
        \hline
        & Teacher & Base & Tiny \\
        \hline
        Hidden Dim & 768 & 768 & 312 \\
        Intermediate Size & 3072 & 3072 & 1200 \\
        Layers & 12 & 6 & 4 \\
        Attention Heads & 12 & 12 & 12 \\
        Max Positions & 512 & 512 & 512 \\
        \hline
    \end{tabular}
    \caption{Student vs Teacher hyperparameters. We reduce the hidden dimensionality, feedforward intermediate size, and the number of layers. The number of attention heads and max number of positions (tokens) are kept the same.}
    \label{tab:distillation_parameters}
\end{table}

\subsection{Downstream Finetuning}
To measure the performance of the distilled model on downstream tasks, we finetune on several benchmarks in Tagalog:

\begin{itemize}
    \item TLUnified NER \cite{miranda2023developing} -- NER classification dataset developed using the TLUnified \cite{cruz-cheng-2022-improving} corpus.
    \item Hatespeech Filipino \cite{cabasag2019hate} -- a text classification dataset on hatespeech mined from election tweets in Tagalog.
    \item NewsPH NLI \cite{cruz2021exploiting} -- an entailment dataset created using news articles in Tagalog.
\end{itemize}

We measure accuracy for the hate speech classification and NLI tasks and measure F1 for the NER task. We compare the performance of our models with mBERT (as the teacher), Tagalog-RoBERTa \cite{cruz-cheng-2022-improving} (to compare against a full model trained on Tagalog), DistilmBERT \cite{sanh2020distilbertdistilledversionbert} (a full distilled version of mBERT retaining all the languages supported), and from-scratch training (where a blank model is directly tuned on the downstream task).



\begin{table*}[]
\centering
\begin{tabular}{l|ll|ll|ll|l|}
\cline{2-8}
                                                 & \multicolumn{2}{c|}{\textbf{TLUnified NER}} & \multicolumn{2}{c|}{\textbf{Hatespeech}} & \multicolumn{2}{c|}{\textbf{NewsPH NLI}} & \multicolumn{1}{c|}{Avg.} \\ \cline{2-7} 
                                                 & F1                & Runtime                & Accuracy                & Runtime                & Accuracy            & Runtime  & Speedup \\ \hline
\multicolumn{1}{|l|}{From Scratch}               & 0.4818                       & 71s                      & 0.7382                       & 617s                      & 0.5392                   & 25819s  &                \\
\multicolumn{1}{|l|}{Tagalog RoBERTa}            & 0.8939                       & 66s                      & 0.7767                       & 606s                      & 0.9406                   & 25798        &          \\
\multicolumn{1}{|l|}{mBERT}                      & 0.8925                       & 70s                      & 0.7543                       & 618s                      & 0.9318                   & 25811s      &            \\
\multicolumn{1}{|l|}{DistilmBERT}                 & 0.8818                       & 44s                      & 0.7372                       & 366s                      & 0.9172                   & 15316s       &  1.68x         \\
\multicolumn{1}{|l|}{dBERT Base (Ours)}         & 0.8074                       & 44s                      & 0.7729                       & 309s                      & 0.9188                   & 13006s          & 1.97x       \\
\multicolumn{1}{|l|}{dBERT Tiny (Ours)}         & 0.6085                      & 31s                     & 0.7261                       & 107s                      & 0.8328                   & 4917s       &  5.23x         \\ \hline
\end{tabular}
\caption{Main Results. Accuracy refers to evaluation accuracy on the test set. Runtime refers to the total amount of time (in seconds) that it takes to finetune on the task dataset (rounded down). Avg. Speed refers to the factor by which the distilled models are faster compared to mBERT (averaged across the three tasks).}
\label{tab:main-results}
\end{table*}

\section{Results and Discussion}
A summary of the results can be found on Table \ref{tab:main-results}.

We can see that our models perform strongly across the three benchmark tasks. For the hate speech classification and NLI tasks, our dBERT Base model outperforms its teacher mBERT as well as the distilled DistilmBERT version with an almost 2x speedup in terms of training time. This shows that the method, albeit simple, works well to produce general-use transformers for these tasks. Performance lags slightly behind on NER, which we assume is a harder task for an extracted model as there are a lot of named entities in the vocabulary from other languages that are not completely removed and present a significant amount of negative interference. We investigate these behaviors further in ablations.

The dBERT Tiny variant showed strong results that came close to the baselines on hate speech classification but lags behind the other models in all other tasks. We hypothesize that this is due to the size of the model not having enough capacity to fully capture the teacher's representation of the target language given that the source representation space is extremely large due to the presence of other languages.

Unsurprisingly, RoBERTa Tagalog performs the best in all three tasks given that it is a full-sized BERT-type model that is trained solely in Tagalog. The mBERT and DistilmBERT models are likewise strong performers but are much slower during training than the dBERT models which has a significant impact on research in low-resource languages where computing is often scarce.

Overall, this provides empirical evidence that distilling a general-purpose transformer from a larger MMT yields robust results despite the method's relative simplicity.

\subsection{Can we outperform the teacher with less training data?}

\begin{table}[]
    \centering
    \begin{tabular}{|c|c|c|}
        \hline
        \textbf{Model} & \textbf{Accuracy} & \textbf{Perf. Diff.} \\
        \hline 
        dBERT @100\% & 0.7729 & +0.0186\\
        dBERT @80\% & 0.7200 & -0.0343 \\
        dBERT @50\% & 0.7108 & -0.0435 \\
        mBERT & 0.7543 &  \\
        \hline 
    \end{tabular}
    \caption{Ablation on the amount of training data used for distillation. Data size refers to how much training data is retained. Accuracy represents accuracy on the test set of Hatespeech Filipino. Perf. Diff. refers to the difference in the performance of the finetuned distilled model against mBERT's finetuned performance on Hatespeech Filipino.}
    \label{tab:lessdata}
\end{table}

One surprising result from the benchmarking is the fact that the student model \texttt{dBERT Base} outperforms its teacher \texttt{mBERT} on hate speech classification by 1.86\% in accuracy. This suggests that a smaller dataset may be as-effective for isolating performance for one language in an MMT as opposed to using a larger one. To further investigate this, we distill more versions of \texttt{dBERT Base} using 80\% and 50\% of the original training data and re-run the experiments for Hatespeech classification. A summary of the results can be found on Table \ref{tab:lessdata}.

We see that when reducing the training data used for distillation, the performance starts to be impacted but not by a significant margin. The original mBERT model only outperforms dBERT @80\% training data by around 3.43\% accuracy on hate speech classification. Once we go down to half the training data, the original only outperforms the student model by 4.35\% -- a sub 1\% degradation in performance! We hypothesize that this is connected to the amount of pretraining data used for the target language in the original MMT. The more robust the MMT's performance is in the target language, the less data might be needed to retain that performance post-distillation.

\subsection{Can we improve the student by properly conditioning the teacher?}


\begin{table}[]
    \centering
    \begin{tabular}{|c|c|c|}
        \hline
        \textbf{Model} & \textbf{F1} & \textbf{Perf. Diff.} \\
        \hline 
        dBERT & 0.8074 & -0.0851\\
        dBERT Conditioned & 0.7587 & -0.1338 \\
        \hline
        mBERT & 0.8925 &  \\
        mBERT Conditioned & 0.8900 & -0.0025 \\
        \hline 
    \end{tabular}
    \caption{Ablation on teacher conditioning. Perf. Diff. refers to the difference in the performance of the finetuned distilled models against mBERT's finetuned performance on TLUnified NER.}
    \label{tab:condition}
\end{table}

In our experiments, the NER results are lackluster when compared against \texttt{DistilmBERT}, which was a distilled version of the original \texttt{mBERT}. We assume that this is because the teacher model is not conditioned properly on the target language and experiences some form of negative transfer during the distillation process as the source representation space is very large. To curb this effect, we experiment with first conditioning the teacher on the training dataset by finetuning using masked language modeling \textit{before} performing distillation. We then finetune on the NER downstream task and evaluate after to compare performance. A summary of the results can be found in Table \ref{tab:condition}.

In the initial results, a conditioned mBERT model experiences very minimal performance degradation when finetuned on MLM prior to distillation by a factor of 0.0025 F1. Once we distill, we find that a student distilled from a conditioned teacher performs significantly worse than without teacher conditioning. We hypothesize that the downstream performance suffers because there is some negative interference occurring in the teacher model during conditioning -- a consequence of having a majority of its parameters being dedicated for languages other than the target language we want -- and this creates further instability during distillation to the student.

This suggests that further conditioning of the teacher to the target language may not be necessary for extracting a language-specific model.

\subsection{What if we initialize the student weights from the teacher?}

\begin{table}[]
    \centering
    \begin{tabular}{|c|c|c|}
        \hline
        \textbf{Model} & \textbf{F1} & \textbf{Perf. Diff.} \\
        \hline 
        dBERT & 0.8074 & -0.0851\\
        dBERT Init & 0.7597 & -0.1330 \\
        dBERT Init+Freeze & 0.7659 & -0.1266 \\
        \hline
        mBERT & 0.8925 &  \\
        \hline 
    \end{tabular}
    \caption{Ablation on weight initialization. Perf. Diff. refers to the difference in the performance of the finetuned distilled models against mBERT's finetuned performance on TLUnified NER.}
    \label{tab:initialization}
\end{table}

In this work, we aim to extract general-use language-specific models from large MMTs in the most straightforward way possible, which is why we originally opted to not do any weight initialization and layer copying tricks commonly found in most knowledge distillation works~\cite{jiao2020tinybertdistillingbertnatural}. However, it will be useful to see how much of a contribution weight initialization is in comparison to our method. For this ablation, we perform the simplest initialization commonly used -- copying the embedding weights of the teacher -- and then freezing them before beginning distillation. Like the previous ablation, we evaluate on the NER downstream task to compare performance with our baselines. A summary of the results can be found in Table \ref{tab:initialization}.

We see that interestingly, the student model performs worse when the embedding layer is initialized from the teacher weights by a factor of -0.1330 F1 score. Freezing the embedding layer while performing distillation does not inhibit the performance loss significantly -- the model now performs 0.1266 F1 worse than the original dBERT model without initialization.

While embedding layer initialization is often useful for retaining teacher knowledge when distilling multilingual models \cite{sanh2020distilbertdistilledversionbert}, we can see some empirical evidence that it might not be as useful in cases where we do not want to recapture the entirety of the original embedding space. For extracting single-language models from multilingual models, it may be useful to not copy the embeddings at all.

\section{Related Work}
Knowledge distillation is an established tool in modern NLP research, especially after the release of BERT in 2018. Most works such as DistilBERT and TinyBERT \cite{jiao2020tinybertdistillingbertnatural} aim to distill the full model while retaining all languages that may be incorporated in the original training data. These models perform well across a number of cross-lingual benchmarks such as XNLI \cite{conneau2018xnli}, but represent a challenge in real-world use especially for low-resource languages.

Recent works have begun to use knowledge distillation for smaller, targeted use-case models. \citet{wibowo2024privileged} explores student initializations to improve task-based performance with minimal training needed, and \citet{ansell-etal-2023-distilling} distills smaller models for the goal of efficiently producing stronger task-based models via further distillation. However, most of these works focus directly on the end task, instead of creating a general-use case student model that is targeted for one language specifically.

\section{Future Work}
The current method provides a strong way to distill a language-specific general-use model from a much larger MMT, while being flexible enough to function as the base for more targeted tasks. For future work, the following may be explored as an augmentation to the current method:

\noindent \textbf{Extrapolating to an Unseen Language} -- Much like in BLOOM+1 \cite{yong-etal-2023-bloom}, we could explore teacher conditioning to add an unseen language to an existing language model.

\noindent \textbf{General Purpose LLMs} -- Moving beyond small pretrained models, we can explore the use of the same method for general purpose multilingual LLMs such as Aya \cite{ustun2024aya} and BLOOMZ \cite{muennighoff2022crosslingual} to see if we can transfer learned instruction-following performance on a language-specific student model.

\section{Conclusion}
In this work, we present an extremely simple method of extracting general-use language-specific transformers from pretrained MMTs that retain the robust performance of the original teacher models. These models and the process of obtaining them are both ideal for research in low-resource languages as both the compute resources and the data available for researchers in these areas are often very scarce. For future work, we present a number of augmentations that can be explored from this relatively simple method, such as unseen language extrapolation, and extension to large language models.

\section*{Limitations}
While we provide good empirical results, we acknowledge a number of limitations in our work, mostly due to a lack of compute resources. We study only one MMT -- mBERT -- to simplify the study. In future work, we aim to have a more diverse set of MMTs to test the method on. We also only limit the study to Tagalog as a case study. For future work, we aim to test the method on a wider variety of low-resource languages, as well as using a benchmark high-resource langauge to compare ablations against. Additionally, our distillation step is quick (three epochs) due to the size of the training dataset and limitations in compute. For future work, we aim to identify the relationship between the size of the training dataset, size of the target language in the pretraining dataset, and the length of distillation.

\bibliography{custom}


\end{document}